\documentclass[sigconf]{acmart}

\AtBeginDocument{%
  \providecommand\BibTeX{{%
    \normalfont B\kern-0.5em{\scshape i\kern-0.25em b}\kern-0.8em\TeX}}}

\copyrightyear{2020}
\acmYear{2020}
\setcopyright{acmcopyright}\acmConference[MM '20]{Proceedings of the 28th ACM International Conference on Multimedia}{October 12--16, 2020}{Seattle, WA, USA}
\acmBooktitle{Proceedings of the 28th ACM International Conference on Multimedia (MM '20), October 12--16, 2020, Seattle, WA, USA}
\acmPrice{15.00}
\acmDOI{10.1145/3394171.3413863}
\acmISBN{978-1-4503-7988-5/20/10}

\usepackage{graphicx}
\usepackage{comment}
\usepackage{amsmath,amssymb} % define this before the line numbering.
\usepackage{color}

\usepackage{dsfont}
\usepackage[ruled,vlined]{algorithm2e}
\usepackage{array}
\newcolumntype{L}[1]{>{\raggedright\arraybackslash}m{#1}}
\newcolumntype{C}[1]{>{\centering\arraybackslash}p{#1}}

\begin{document}

\fancyhead{}

\title{Traffic-Aware Multi-Camera Tracking of Vehicles Based on ReID and Camera Link Model}


\author{Hung-Min Hsu}
\email{hmhsu@uw.edu}
\affiliation{%
  \institution{University of Washington}
  \city{Seattle}
  \state{Washington}
  \postcode{98195}
}

\author{Yizhou Wang}
\email{ywang26@uw.edu}
\affiliation{%
  \institution{University of Washington}
  \city{Seattle}
  \state{Washington}
  \postcode{98195}
}

\author{Jenq-Neng Hwang}
\email{hwang@uw.edu}
\affiliation{%
  \institution{University of Washington}
  \city{Seattle}
  \state{Washington}
  \postcode{98195}
}

%\author{Kwang-Ju Kim}
%\email{kwangju@etri.re.kr}
%\affiliation{%
%  \institution{Electronics and Telecommunications Research Institute}
%  \city{Daegu}
%  \country{Korea}
%   \postcode{98195}
%}

\renewcommand{\shortauthors}{Hsu, et al.}

\begin{abstract}
Multi-target multi-camera tracking (MTMCT), i.e., tracking multiple targets across multiple cameras, is a crucial technique for smart city applications. In this paper, we propose an effective and reliable MTMCT framework for vehicles, which consists of a traffic-aware single camera tracking (TSCT) algorithm, a trajectory-based camera link model (CLM) for vehicle re-identification (ReID), and a hierarchical clustering algorithm to obtain the cross camera vehicle trajectories. First, the TSCT, which jointly considers vehicle appearance, geometric features, and some common traffic scenarios, is proposed to track the vehicles in each camera separately. Second, the trajectory-based CLM is adopted to facilitate the relationship between each pair of adjacently connected cameras and add spatio-temporal constraints for the subsequent vehicle ReID with temporal attention. Third, the hierarchical clustering algorithm is used to merge the vehicle trajectories among all the cameras to obtain the final MTMCT results. Our proposed MTMCT is evaluated on the CityFlow dataset and achieves a new state-of-the-art performance with IDF1 of 74.93\%. 
\end{abstract}





%     \includegraphics[width=\textwidth]{figures/framework.png}
%     \caption{The illustration of our MTMCT framework. First, single camera tracker TNT append by our TSCT is utilized to obtain SCT results for each camera. Second, CLM with entry/exit zones and transition times are automatically generated. Third, vehicle ReID with temporal attention is implemented on the solution space from the CLM. Finally, the hierarchical clustering is involved for the final MTMCT results.}
%     \label{fig:framework}

\maketitle

\section{Introduction}
\label{sec:introduction}

\begin{figure*}
    \includegraphics[width=0.8\textwidth]{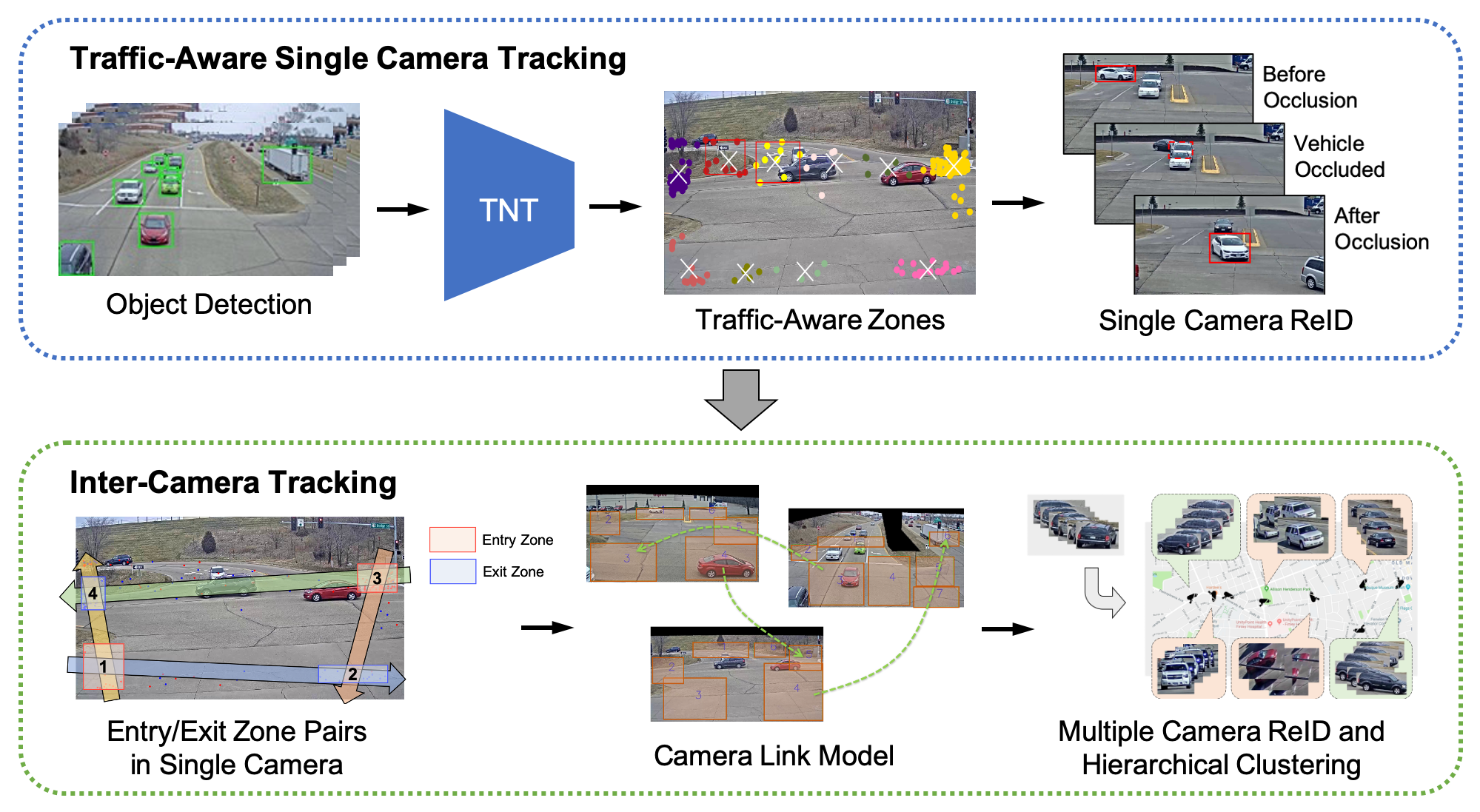}
    \caption{The illustration of our MTMCT framework. First, the traffic-aware single camera tracking (TSCT) is utilized to obtain SCT results for each camera. Second, CLMs with entry/exit zones and transition times are automatically generated. Third, vehicle ReID with temporal attention is implemented on the solution space from the CLM. Finally, the hierarchical clustering is involved for the final MTMCT results.}
    \label{fig:framework}
\end{figure*}

Due to the exponential growth of the deployed surveillance cameras with networking supports, the opportunity to take advantage of the rich information from the multi-camera systems is immense. Among the techniques for multi-camera systems, multi-target multi-camera tracking (MTMCT) is important for traffic flow optimization and anomaly detection. Basically, MTMCT consists of two sub-tasks: 1) Single camera tracking (SCT): Detection and tracking of the objects within each single camera; 2) Inter-camera tracking (ICT): Associations of the object trajectories across different cameras. Overall, MTMCT is aimed to obtain the trajectories of every object in the scene through all the cameras in the system.

However, MTMCT is a very challenging task due to unreliable object detection, heavy occlusion, low resolution, and varying lighting and viewing-perspective conditions. Recently, many proposed works for MTMCT take into account target motion and human pose. Among these works, most of them are person based MTMCT, which only considers humans as the tracking targets. For vehicles, this task becomes more challenging because: 1) vehicles may stop for a long time at the traffic signs and continually be occluded among each other, which makes occlusion even more severe; 2) inter-class similarity for vehicles is higher because there may exist many different identities with the similar appearance. 

To solve these two aforementioned challenges, we propose a novel MTMCT framework, which mainly consists of two innovations, i.e., traffic-aware single camera tracking (TSCT) and the trajectory-based camera link model (CLM). 

First, TSCT is proposed to handle the long-term occlusions created in the traffic scenarios.
Usually, there will be a large number of isolated and fragmented vehicle trajectories, created from a single camera multi-target tracker, in the center of the frames where vehicles do not enter or exit the camera's field of view (FoV).
For example, when a vehicle stops in front of a red traffic light, it can be partially or even fully occluded in the camera's FoV for a long time. We would like to call this kind of zone as a traffic-aware zone. 
Here, we use the TrackletNet tracker (TNT) \cite{wang2019exploit}, which is a superior SCT method in intelligent transportation system application \cite{hsu2019multi,wang2019anomaly,wang2019monocular}, as our single camera tracker.
According to this condition, TSCT is proposed to find out the traffic-aware zones, where this kind of occlusion happens, by clustering the start and end nodes of all the resulting trajectories from the TNT. Vehicle Re-ID in the single camera is then implemented for these traffic-aware zones to connect these disconnected trajectories created in the traffic scenarios. 

Second, facing higher inter-class appearance similarity of distinct vehicles, trajectory-based CLM is further proposed to impose spatio-temporal constraints and reduce solution search space for the cross camera ReID. For two different vehicles with a very similar appearance, it is nearly impossible to re-identify them using a typical ReID method. However, taking advantage of the spatial and temporal constraints between a pair of adjacently connected cameras, we can easily filter out the vehicles that are not likely to appear in a certain camera at a certain timestamp. We define these constraints, including the vehicle entry/exit zones and the transition times, as the CLM. Using the CLM automatically generated from the training data, cross camera vehicle ReID becomes much more accurate and efficient. 

Finally, a hierarchical clustering algorithm, based on the Euclidean distance between the feature space of different trajectories, is used to merge the trajectories among all the cameras to obtain the final MTMCT results. 

Overall, the framework of our proposed MTMCT is shown in Fig.~\ref{fig:framework}. First, we use the TNT \cite{wang2019exploit} to obtain SCT results for each camera. Based on the results from the TNT, our proposed TSCT can generate the traffic-aware zones and perform vehicle ReID within each camera to solve the occlusion problem. We then analyze the vehicle trajectories in the training dataset and automatically generate the CLM, i.e., the entry/exit zones and the transition times, for each camera. After that, vehicle ReID with temporal attention is implemented on the solution space from the CLM. Finally, the hierarchical clustering is introduced for the final MTMCT results. To summarize, we claim the following contributions:
\begin{itemize}
    \item Propose a new MTMCT framework specifically designed for vehicles.
    \item Utilize a novel  TSCT strategy to improve vehicle SCT results, considering common traffic scenarios.
    \item Create a trajectory-based CLM generation method that adds spatio-temporal constraints and reduces the solution space for cross camera vehicle ReID.
    \item Achieve a new state-of-the-art performance on the CityFlow dataset.
\end{itemize}

The rest of the paper is organized as follows. In Section~\ref{sec:related_works}, we provide an overview of the related works. Our proposed framework for MTMCT of vehicles is introduced in Section~\ref{sec:mtmct}. Section~\ref{sec:experiments} presents our experimental results of the proposed MTMCT framework on the CityFlow dataset \cite{tang2019cityflow}. Finally, the conclusion is drawn in Section~\ref{sec:conclusion}.

\section{Related Works}
\label{sec:related_works}

%In the past few years, person based ReID and MTMCT schemes have attracted more and more attention \cite{gray2008viewpoint,hirzer2011person,cheng2011custom,li2012human,li2014deepreid,zheng2015scalable,zheng2016mars,ristani2016performance,chen2016equalized,zheng2017unlabeled,wei2018person,wu2018exploit}. Moreover, some works are also concerned about vehicle based ReID \cite{liu2016deep,yan2017exploiting} due to smart city related applications. Different from person ReID, there are two main challenges in vehicle ReID, which are small inter-class variability (e.g., different cars with similar car appearance) and large intra-class variability (e.g., the same car with different lighting, distance, and/or viewing angles from the same or different cameras).% 
% In this section, we review the related works from three MTMCT components: single camera tracking, appearance feature based vehicle ReID, and camera link model.

\paragraph{Single camera multi-object tracking. } 
Tracking-by-detection based multi-object tracking approaches \cite{cai2020ia,zhang2020lifts} are the most popular schemes for single camera tracking (SCT). % The procedure of the tracking-by-detection approach is to associate detections across frames. If the detection is not unreliable or occlusions occur, the tracking-by-detection approach needs to estimate object locations from tracking results. % 
Many approaches define SCT as a graph optimization problem \cite{kumar2014multiple,tang2016multi,tang2017multiple,wen2014multiple,choi2015near,tang2015subgraph,tang2018single}, i.e., each detected object is represented as a vertex to form a graph with edges denoting the affinity (similarity) between two detections in two image frames, while in some works \cite{wen2014multiple,choi2015near,tang2015subgraph,tang2018single}, a  vertex denotes a tracklet formed from some association rules, with the edges denoting the affinity between two disconnected tracklets. 
%However, there are some issues in detection-based graph models for SCT. For example, graph models are assumed to be conditionally independent of the vertices. But detections are not conditionally independent cross frames in SCT, temporal information can be used to improve the performance of SCT. Moreover, the graph is represented as a high-dimensional affinity matrix so that it is difficult to seek the global minimum solution in the graph clustering optimization process. In contrast to the detection-based graph models, the tracklet based graph model utilizes the information from a short trajectory to facilitate estimating with better robustness the relationship between vertices.%
The appearance feature plays a vital role in the tracking-by-detection framework. Many different types of appearance features are applied to deal with SCT. For example, CNN-based features have been widely used for ReID tasks \cite{zhang2017multi,tang2017multiple,zhong2017re,ristani2018features}, where metric learning is exploited to train the CNN-based features \cite{ristani2018features}. %A re-ranking technique \cite{zhang2017multi,zhong2017re} is adopted in calculating the feature similarity. Histogram-based features, such as color histograms, HOG, and LBP, are also adopted as appearance features. On the other hand, temporal features, such as the location, size, and motion of bounding boxes along the time, are also commonly used in the SCT. %
Furthermore, appearance features and temporal features can be combined \cite{zhang2017multi,tang2018single}  to achieve better performance. However, none of these methods can fully handle the long occlusion caused by the different traffic scenarios. 

\paragraph{Camera link model. }
A camera link model (CLM) consists of camera link information and transition time distribution of a pair of adjacently connected cameras, i.e., CLM takes spatio-temporal constraints into consideration. In MTMCT, the CLM can be used as effective constraints to reduce the search space of matching so as to improve the performance of MTMCT, as evidenced in some works \cite{lee2015combined,tang2018single,tang2019cityflow,hsu2019multi}. For example, Lee et al.~\cite{lee2015combined} use the CLM  to estimate bidirectional transition time distribution in an unsupervised scheme for MTMCT. Tang et al.~\cite{tang2018single,tang2019cityflow} and Hsu et al.~\cite{hsu2019multi} also use the car speed to generate the transition time distribution for each connected pair of adjacently connected cameras. Therefore, the accuracy of MTMCT can be significantly improved by CLM by reducing the candidate set of matching. In this paper, we systematically generate the CLMs to assist the ReID in the MTMCT tasks.

\paragraph{Appearance feature based vehicle ReID} 
Vehicle ReID has attracted more research efforts in the past few years. VeRi-776 \cite{liu2016deep} and VeRi-Wild \cite{lou2019veri} are the most widely used benchmarks,  which provide not only the high quality annotations but also the camera deployment geometry. In ReID, there are two types of appearance feature extraction methods. One is traditional handcrafted methods, such as SURF \cite{bay2006surf} and ORB \cite{rublee2011orb}. The other is CNN-based methods \cite{zhao2016spectral,wu2018light}, which have been proven to achieve better performance than traditional handcrafted methods in the past years. Thus, recent methods \cite{liu2016deep,liu2016deep1} focus on learning an embedding model that maps the samples into an embedding space. Liu et al.~\cite{liu2016deep} propose a mixed difference network using a vehicle model and ID information to construct more robust feature embedding. Liu et al.~\cite{liu2016deep1} propose a ``PROVID'' ReID model by using visual feature, license plate and spatio-temporal information for the vehicle ReID task. Shen et al.~\cite{shen2017learning} incorporate complex spatio-temporal information for effectively regularizing the ReID results by a two-stage framework. Based on a multi-view inference scheme, Zhou et al.~\cite{zhou2018aware} generate a global-view feature representation to improve the vehicle ReID. Huang et al.~\cite{huang2019multi} utilize car key points to train orientation-based embedding for vehicle ReID. Tang et al.~\cite{tang2019pamtri} propose a pose-aware multi-task re-identification (PAMTRI) framework by explicitly reasoning about vehicle pose, shape, color, and types. In this work, we propose a trajectory-based CLM to enhance the performance of vehicle ReID.

\section{Proposed Method}
\label{sec:mtmct}

There are four steps in our proposed method, as illustrated in Fig.~\ref{fig:framework}. First, we apply TSCT after a single camera tracker TNT, as presented in Section~\ref{subsec:tsct}. Then, the camera links are established based on the generated entry/exit zones, as illustrated in Section~\ref{subsec:clm}. Moreover, we utilize a vehicle ReID method by combining the temporal attention and batch sampling for inter-camera tracking, as explained in Section~\ref{subsec:reid}. Finally, we use the hierarchical clustering to merge the tracklets into object trajectories, as described in Section~\ref{subsec:hc}. 

\subsection{Traffic-Aware Single Camera Tracking (TSCT)}
\label{subsec:tsct}

In our MTMCT framework, the first step is SCT, where we implement the TrackletNet Tracker (TNT) \cite{wang2019exploit}. A TNT is a tracklet-based graph clustering approach, where the vertices are generated based on detection association via CNN-based appearance feature and the intersection-over-union (IOU) in the two consecutive frames. While, the edge weights are estimated by a TrackletNet, which is a Siamese neural network trained to predict the likelihood of the two tracklets belonging to the same object. After the tracklet-based graph is constructed, graph clustering \cite{tang2018single} is applied to merge the tracklets of the same vehicle into one single trajectory.

After using TNT to generate the SCT results, we observe that there are some vehicle ID switches due to some \emph{isolated trajectories}, which usually appear while vehicles are waiting for a traffic light. Therefore, we propose TSCT to deal with this issue by generating the traffic-aware zones and performing single camera ReID to reconnect the isolated trajectories caused by the traffic scenarios. 

First of all, all the traffic-aware zones within each camera are detected in an unsupervised manner based on the MeanShift clustering algorithm \cite{comaniciu2002mean} applied on the collected entry/exit measurements. 
The procedure of traffic-aware zone generation is as follows: 1) use the first and the last positions of all the trajectories from the SCT results as zone nodes; 2) cluster the zone nodes into different groups by apply the MeanShift algorithm. 

\begin{figure*}[t]
    \centering
    \includegraphics[width=0.75\linewidth]{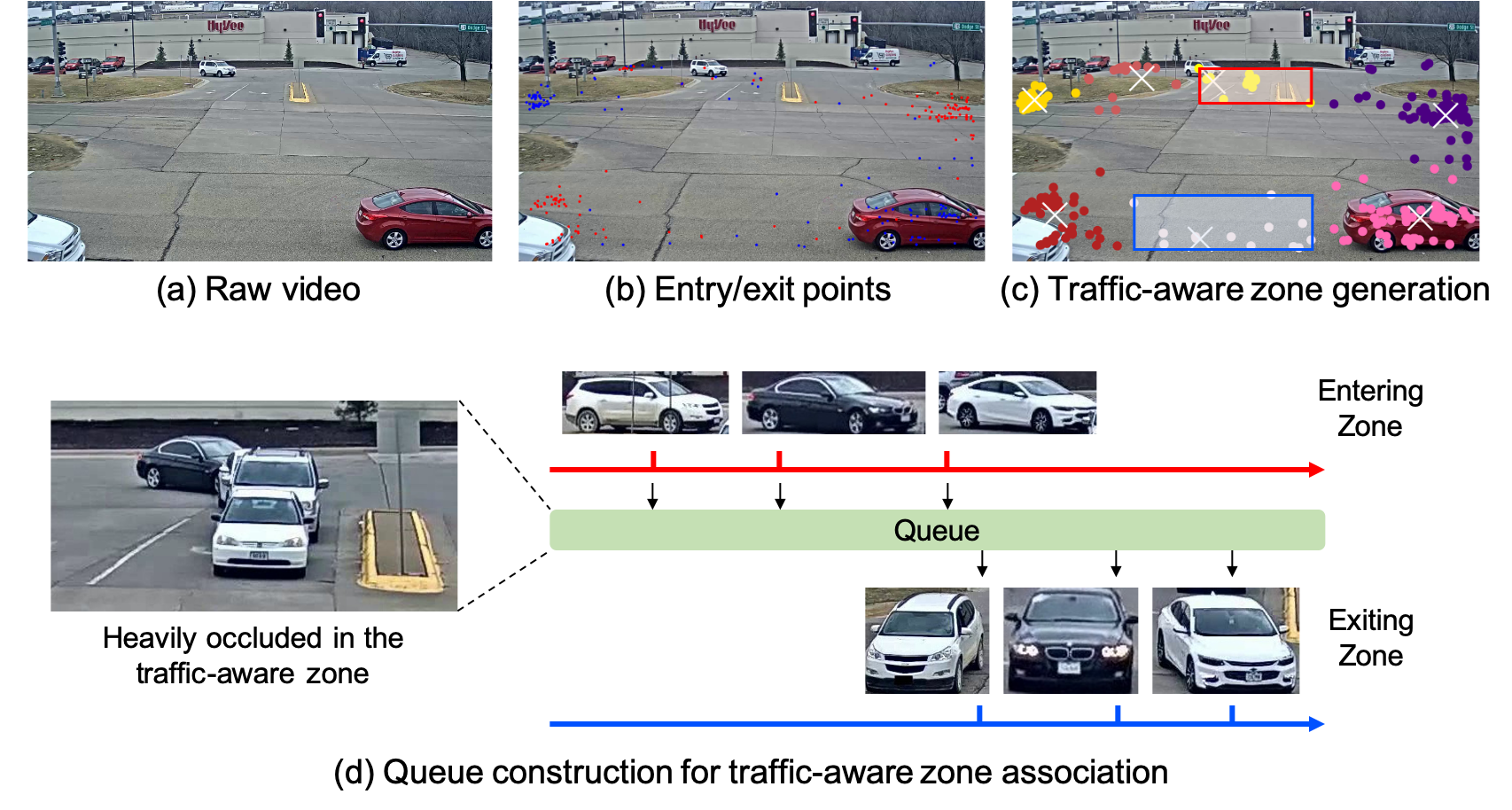}
    \caption{Traffic-aware zone generation. (a) Raw video from a surveillance camera. (b) Start/end points of all trajectories from SCT (red: entry; blue: exit). (c) The clustering result for traffic-aware zone generation from Meanshift (red: stop for traffic light; blue: truncated vehicles). (d) A queue for isolated trajectories is maintained to keep the ordering and appearance features for following single camera ReID.}
    \label{fig:traffic_aware_zone}
\end{figure*}

After the MeanShift, we define a rectangular zone for each group to bound the inside nodes. We denote $N_{e,k}$ as the number of entry nodes and $N_{x,k}$ as the number of exit nodes in the zone $k$. The traffic-aware density $D_{ta}$ is defined by 
\begin{equation}
    D_{ta} = 1 - \frac{|N_{e,k} - N_{x,k}|}{N_{e,k} + N_{x,k}},
\end{equation}
where $D_{ta}$ needs to be above a threshold $\rho_{ta}$, the zone will be designated as a traffic-aware zone, as shown in Fig.~\ref{fig:traffic_aware_zone}(c).

After the traffic-aware zones are generated, the next step is to reconnect the isolated trajectories in these zones. Here, we build a queue for each zone to store the ordering and appearance features of the vehicles that are interrupted in the traffic-aware zones. If there is a bounding box of an isolated trajectory that suddenly appears in the traffic-aware zone, we select a trajectory from the queue to compute the appearance similarity. The appearance feature is trained by single camera ReID, which is similar to the cross camera ReID, as explained in Section~\ref{subsec:reid}. 
% There are some abnormal trajectories that finished or start in the middle of a frame. If the final detection of a trajectory finishes in the middle of a frame (i.e, not finish in the exit zone), this trajectory is defined as Tr$em$. Those trajectories which start from the middle of a frame (i.e., not start from an entry zone) are defined as Tr$sm$. Calculate the IOU of Tr$em$’s first BBOX and the traffic-aware zones in the camera, if the IOU$et$ over a threshold, the trajectory will be inserted into a queue. After that, if there is a Tr$sm$ that IOU$st$ with the traffic-aware zones is over a threshold, we compute the IOU$es$ between the last detection bounding box of Tr$em$ and the first detection bounding box of Tr$sm$. The IOU$es$ also need to over a threshold then we will estimate the appearance similarity of Tr$em$ and Tr$sm$. The appearance feature is trained by single camera ReID which is similar to the cross camera ReID (explained in Section 3.3). The only difference is that we refer the same car in the same camera as the same identity which means that the same car in the different cameras is treated as a different identity. The illustration is shown in Fig.~\ref{fig:traffic_aware_zone}(d).

\subsection{Trajectory-based Camera Link Model}
\label{subsec:clm}

Our trajectory-based CLM can be divided into three steps: 1) entry/exit zones generation in each single camera; 2) vehicle trajectory classification according to entry-exit zone pairs in each single camera; 3) camera links and transition time estimation across different cameras. These three steps are described below.

\paragraph{Entry/exit zones generation. }
With the available routes and the detected entry/exit zones, we can estimate the relationship between each two zones. For a trajectory-based CLM, the procedure of the entry/exit zone generation is similar to the traffic-aware zone generation. The only difference is to calculate the entry and exit density to determine whether the generated zone is an entry/exit zone or not. The entry and exit densities are defined as $D_e$ and $D_x$, where 
\begin{equation}
    D_e = \frac{N_{e,k}}{N_{e,k} + N_{x,k}}, \ 
    D_x = \frac{N_{x,k}}{N_{e,k} + N_{x,k}}.
\end{equation}
If the density of an entry or exit zone is higher than a threshold $\rho_e$ or $\rho_x$, this zone will be recognized as an entry or exit zone. There may be multiple entry/exit zones within one camera's FoV. 

% So far, we can only obtain the zones of each camera without knowing the zone connection among the cameras. It is important to know the adjacently connected zones because we can use the entry and exit observations from the associated zones to reduce the solution search space, i.e., the number of entry-zone candidates to be matched when a car exits from one camera exit zone, for vehicle ReID to obtain better performance of MTMC.

\begin{figure}[t]
    \centering
    \includegraphics[width=0.83\linewidth]{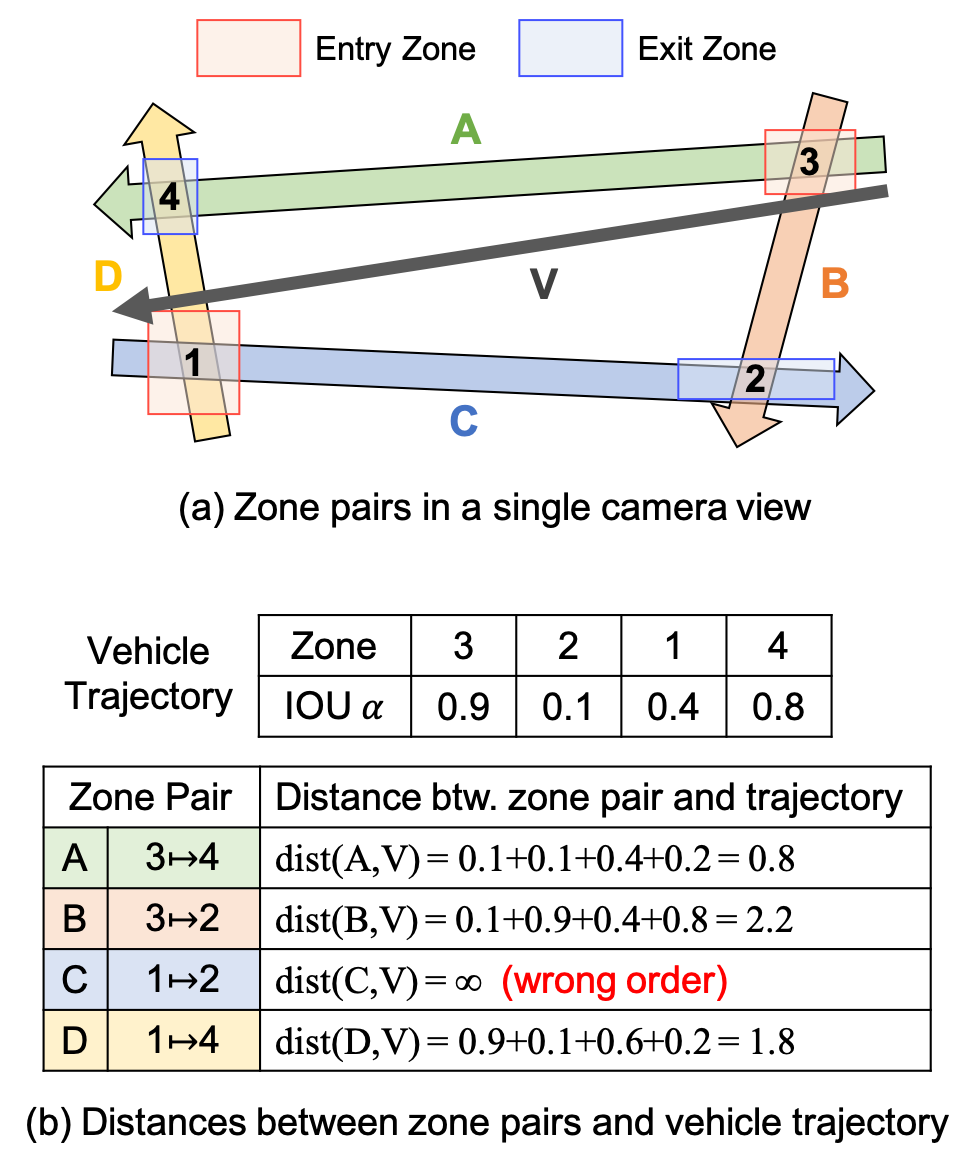}
    \caption{Illustration of distance calculation between a vehicle trajectory (black) and each entry-exit zone pair. There are four zone pairs (A, B, C, D) with two entry and two exit zones in current frame. The vehicle pass through the four zones with certain IOU in the upper table in (b). The distances are calculated according to Eq.~\ref{eq:dist_vt_pairs}. 
    As a result, this vehicle trajectory is classified to zone pair A, which has the smallest distance.}
    \label{fig:zone_pairs}
\end{figure}

\paragraph{Vehicle trajectory classification by zone pairs. }
After the zones are generated, all of the trajectories need to be classified according to entry-exit zone pairs. In our model, an entry-exit zone pair is used to uniquely describe a trajectory, which is called a zone pair trajectory, as shown in Fig.~\ref{fig:zone_pairs}(a). In this paper, we generate all possible camera links by using the training data of MTMCT to systematically generate the camera links instead of human labeling. In most cases, the straight and the right-turn trajectories can be described using different zone pairs being passed through. Sometimes the trajectory may not pass through the corresponding zone pair perfectly due to the viewing angle of the camera, i.e., also passes through the neighboring zones associated with the adjacently connected zone pair. In this case, measuring the distance (mismatch) between a tracked vehicle and a zone-pair trajectory is necessary. The distance can be calculated as 
\begin{equation}
    \text{dist}(P, V) = \sum_{z \in P \cup V} |\mathds{1}(z \in P) - \alpha_z|,
    \label{eq:dist_vt_pairs}
\end{equation}
where $P$ denotes the zone-pair trajectory and $V$ is the actual zones gone through by the tracked vehicle, $\alpha_z$ represents the overlapping ratio of the vehicle to zone $z$, i.e., the overlapping area divided by the vehicle bounding box area. Furthermore, the order of the zones in the zone pair and the tracked vehicle is also considered. Once the order in the tracked vehicle conflicts with the zone pair, the distance between a tracked vehicle and a zone-pair trajectory is set to infinity. Finally, the closest zone-pair trajectory with minimum distance is assigned by comparing the tracked vehicle with all the possible zone-pair trajectories within the camera. An example of distance calculation is shown in Fig.~\ref{fig:zone_pairs}(b). 

% camera link model
\paragraph{Camera link model construction. }
Given the locations of the cameras, we can obtain the routing information provided by the training data. The routing information contains all the links between every two adjacently connected cameras. If there is one link that connects two cameras without passing through another camera, we define them as a camera link. In other words, if all the routes from the training data between two cameras passed by at least one other camera, this link should not exist in our camera link model. 

\begin{figure}
    \centering
    \includegraphics[width=0.70\linewidth]{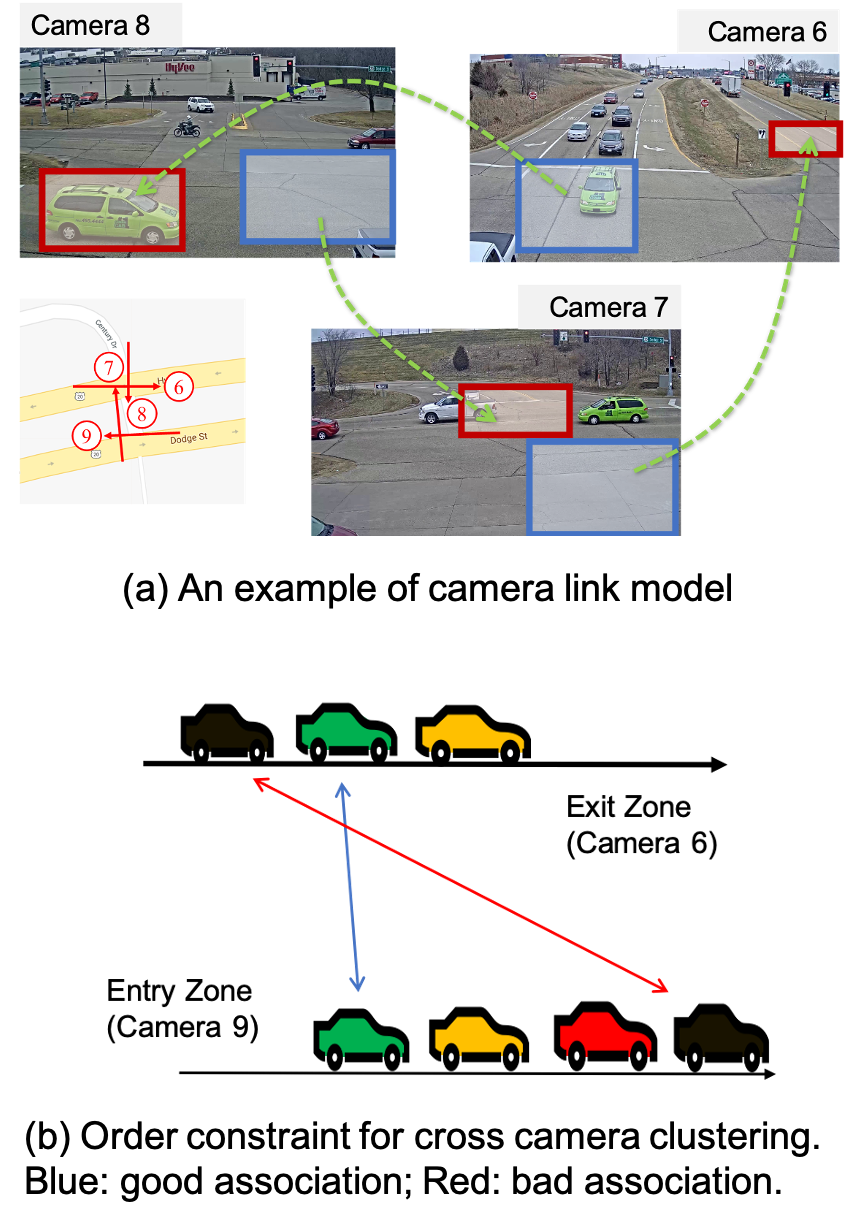}
    \caption{Examples of camera link model and the order constraint. There are three cameras with overlapping FoV, the blue/red bounding boxes represent the exit/entry zones, respectively. According to (a), the location of the exit zone in the Camera 6 is the same as the location of the entry zone of Camera 8, which means  the vehicle exits the Camera 6 will immediately appear in the Camera 8. Similarly, Camera 7 connects to Camera 6 and Camera 8 links to Camera 7.}
    \label{fig:transition_time}
\end{figure}

%The last step of the trajectory-based camera link model generation is to estimate the transition time of camera pairs.%
The camera link and the corresponding transition time of each camera pair can be defined as $\mathcal{T} = (C^{s}, C^{d})$, where $C^{s} = \{ P^{s}_{i}\}_{i=1}^m$ is the zone pair trajectories set in the source camera and $C^{d} = \{P^{d}_{j}\}_{j=1}^n$ is that set in the destination camera. Each camera pair can have more than one transition time due to the bi-directional traffic. In the camera pair with the overlapping view, $\mathcal{T}$ usually only contains one single zone pair trajectory, while for the non-overlapping view case, $\mathcal{T}$ can involve multiple zone pair trajectories. An example to explain the concept of our camera link model is shown in Fig.~\ref{fig:transition_time}(a). 

\paragraph{Transition time estimation. }
To estimate the transition time of each camera link $\mathcal{T}$, we first define the transition zones $z_{s}$ and $z_{d}$, such that $z_{s} \in P^s_{i}$ ($\forall P^s_{i} \in C^s$) and $z_{d} \in P^d_{j}$ ($\forall P^d_{j} \in C^d$). Then, the transition time can be applied as the temporal constraint for both $C^s$ and $C^d$. Given a camera link of a vehicle trajectory from $P^s$ to $P^d$, i.e., from source camera to destination camera, the transition time is defined as
\begin{equation}
    \Delta t = t^s - t^d,
\end{equation}
where $t^s$ and $t^d$ represent the time of the tracks passing $z^s$ and $z^d$, respectively. We can then obtain a time window $(\Delta t_{\text{min}}, \Delta t_{\text{max}})$ for each camera link $\mathcal{T}$ so that only the tracked vehicle pairs whose transition time are within the time window are considered as valid. Thus, the search space of the ReID can be greatly reduced by using the appropriate time window. Then, we can use the embedding feature from the ReID model, trajectory-based camera link model, and clustering algorithm to produce the global IDs for MTMCT. 

\subsection{Cross Camera ReID} 
\label{subsec:reid}

After TSCT, the SCT results for each camera are available as the input data for MTMCT. Video-based ReID can achieve better performance than image-based ReID since it can take advantage of temporal continuity of a sequence of image frames instead of a single image. Therefore, an MTMCT system usually uses trajectory-level features for ReID. The feature extractor is based on the ResNet-50 \cite{he2016deep}, which is pre-trained on ImageNet, and the appearance feature of a vehicle is a 2048-dim vector output by the fully connected layer.

Once the frame-level features are extracted, we use temporal attention (TA) mechanism  \cite{gao2018revisiting} to aggregate the frame-level features into the clip-level features. After that, we use the average pooling to generate trajectory-level features. In the TA modeling, there are two convolutional networks, one is a spatial convolutional network with 2D convolutions and the other is a temporal convolutional network with 1D convolution. Through training these two networks, an attention vector $v_{att}$ can be obtained to weight the frame-level features $\mathbf{f}_{frame}$, resulting in the clip-level features, where
\begin{equation}
    \mathbf{f}_{clip} = v_{att} \cdot \mathbf{f}_{frame}.
\end{equation}

In terms of network training, metric learning is adopted to enhance this vehicle ReID task. To train the model in a more efficient way, we adopt the batch sample (BS) scheme \cite{kuma2019vehicle} in the triplet generation. Overall, in our final loss function, the BS triplet loss is combined with the cross-entropy loss to jointly exploit distance metric learning and identity classification,
\begin{equation}
    \mathcal{L}_{total} = \lambda_1 \mathcal{L}_{BStri} + \lambda_2 \mathcal{L}_{Xent}.
\end{equation}

The objective of triplet loss is to minimize the feature distance of the same identity and maximize the feature distance of different identity pairs \cite{hermans2017defense}. For the BS triplet loss, the objective is to calculate triplet loss $\mathcal{L}_{BStri} (\theta; \xi)$ in a minibatch defined as
\begin{equation}
    \mathcal{L}_{BStri} (\theta; \xi) = \sum_{b} \sum_{a \in B} l_{triplet}(a),
\end{equation}
where
\begin{equation}
    l_{triplet}(a) = \left[ m + \sum_{p \in P(a)} w_p D_{ap} - \sum_{n \in N(a)} w_n D_{an}\right] _+.
\end{equation}
Here, $w_p$ and $w_n$ are the weightings of positive and negative samples, $m$ represents the margin, $D_{ap}$ and $D_{an}$ are the distances between the anchor sample to the positive sample and negative sample, respectively.

Based on the BS strategy, the weightings of positive and negative samples are defined as follows,
\begin{equation}
\begin{aligned}
    w_p = \mathcal{P}(w_p == \text{multinomial}_{x \in P(a)} \{D_{ax} \}),\\
    w_n = \mathcal{P}(w_n == \text{multinomial}_{x \in N(a)} \{D_{ax} \}),
\end{aligned}
\end{equation}
where $x_p$ and $x_n$ are positive and negative samples, respectively.

The cross-entropy (Xent) loss in the training is defined as
\begin{equation}
    \mathcal{L}_{Xent} = - \frac{1}{N} \sum_{i=1}^{N} q(i) \cdot \log(\text{prob}(i)),
\end{equation}
where $\text{prob}(i)$ is the probability of the probe vehicle belongs to vehicle $i$, $q(i)$ is the ground truth vector of vehicle $i$, and $N$ is the number of vehicles in training data.

\subsection{Hierarchical Clustering}
\label{subsec:hc}

\begin{figure}[t]
    \centering
    \includegraphics[width=0.9\linewidth]{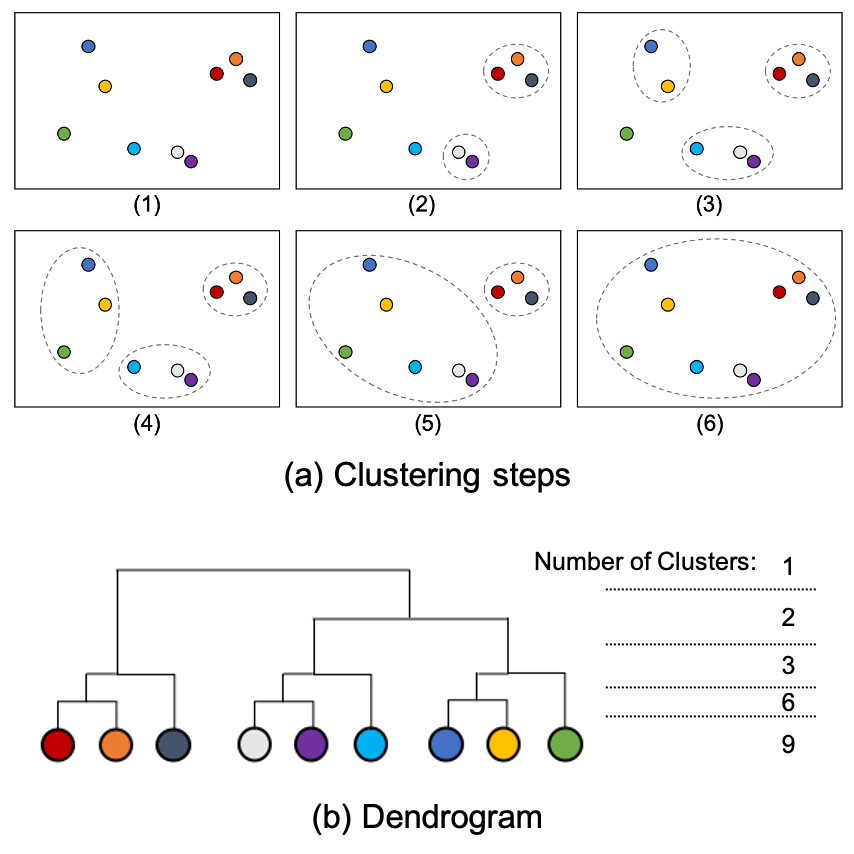}
    \caption{Illustration of hierarchical clustering. An example is shown in (a) for the clustering steps. (b) shows the dendrogram of the example with the number of clusters at different threshold levels.}
    \label{fig:hc}
\end{figure}

\begin{algorithm}[t]
    \SetAlgoLined
    \SetKwInOut{Input}{Input}
    \SetKwInOut{Output}{Output}
    \SetKw{KwFrom}{from}
    \SetKw{KwAnd}{and}
    \Input{Trajectories set $\Xi = \{\xi_n\}$ from all cameras.}
    \Output{Global ID for all trajectories within all cameras.}
     Initialize distance matrix $\mathbf{M}$ between each two trajectories $\mathbf{M} \gets \infty$\;
     \For{trajectory $\xi_i$ \KwFrom $\Xi$}{
      \For{trajectory $\xi_j$ \KwFrom $\Xi$}{
       $\mathbf{M}_{i,j} \gets \text{dist}(\mathbf{f}(\xi_i), \mathbf{f}(\xi_j))$\;
      }
     }
     Flatten and sort the upper triangular part of $\mathbf{M}$ in ascending order: $F \gets sort([\mathbf{M}_{i,j}]_{i<j})$\;
     $iter \gets 0$, $\delta \gets$ distance threshold\;
     \While{$iter <$ \# of iterations}{
      \For{$m_{i,j}$ \KwFrom $F$}{
          \eIf{$m_{i,j} < \delta$ \KwAnd valid order constraint for $\xi_i$, $\xi_j$}{
           Assign the same global ID to $\xi_i$ and $\xi_j$\;
           }{
           $m_{i,j} \gets \infty$\;
           }
      }
      $iter ++$\;
     }
     \caption{Hierarchical Clustering}
\end{algorithm}

We perform data association between a pair of single camera trajectories corresponding to an adjacently connected exit/entry zone for multi-camera trajectories utilizing correlation clustering in each separate short time windows. The time window $(\Delta t_{\text{min}}, \Delta t_{\text{max}})$ is generated from CLM. In each window, there is a weighted graph $\mathcal{G} = (\mathcal{V},\mathcal{E},\mathcal{W})$, where $\mathcal{V}$ represents the single camera trajectory node set, the weight $\mathcal{W}$ of the edge $\mathcal{E}$ represents the corresponding correlation between the nodes. Therefore, MTMCT can be referred as a correlation clustering problem. Then the issue is to partition the node set into subsets. The single camera trajectories of the same identity should belong to the same  subset. Edges in the same subsets accumulate high positive correlations, while edges in the different subsets accumulate high negative correlations. The problem can thus be defined as the following Binary Integer Program (BIP),
\begin{equation}
    \begin{aligned}
    & X^* = \arg \max_{\{x_{i,j}\}} \sum_{(i,j) \in E} w_{i,j} x_{i,j}, \\
    \text{s.t.} \ \  & x_{i,j} + x_{j,k} \leq 1 + x_{i,k}, \ \ \forall i,j,k \in V.
    \end{aligned}
    \label{eq:bip}
\end{equation}
The set $X$ is the set of all possible combinations of assignments to the binary variables $x_{i,j}$. We maximize the summation of  $w_{i,j} x_{i,j}$, which rewards edges that connect the same vehicle's multi-camera trajectories and penalizes edges that link to the different vehicles. If the two multi-camera trajectories $i, j$ are from the same vehicle, then $x_{i,j}$ should be assigned $1$. The constraints in Eq.~\ref{eq:bip} enforce the transitivity in the solution.  

To reduce the computational complexity, we use hierarchical clustering in our method. Fig.~\ref{fig:hc} illustrates how to cluster the single camera trajectories into cross camera trajectories.  Since the search space of ReID is reduced by the transition time constraint, the Rank-$1$ accuracy will be close to $1$. Therefore, we can greedily select the smallest pair-wise distance to merge the tracked vehicles cross cameras. Furthermore, the order between different tracked vehicles can be used as a constraint to further reduce the search space of the ReID. Due to the traffic scenarios or the road conditions, the orders of vehicles should be almost the same. Take Fig.~\ref{fig:transition_time}(b) as an example, we will remove the pairs whose orders conflict with those of previously matched pairs. The process will repeat until there is no valid transition pair, or the minimum distance is larger than a threshold. 
% Given two tracked vehicles $tr_{src1}$ and $tr_{src2}$ in source camera and two tracked vehicles $tr_{src1}$ and $tr_{src2}$, 
% \begin{equation}
%     \text{sign}(t_{src1} - t_{src2}) = \text{sign}(t_{dst1} - t_{dst2}),
% \end{equation}
% i.e., the orders of tracked vehicles in source and destination camera should remain the same. In this case, the search space can be greatly reduced. 

\section{Experiments}
\label{sec:experiments}

\subsection{Dataset and Evaluation}

Our research aims to addressing multi-target multi-camera tracking of vehicles from the video sequences. CityFlow \cite{tang2019cityflow} is the largest and the most representative MTMCT dataset for practical scenarios, which is proposed in CVPR 2019 by Nvidia. To the best of our knowledge, it is the only existing city-scale traffic camera dataset. 
% Other datasets, e.g., VeRi-wild, DukeMTMC, are not really applicable for our scenarios. VeRi-wild is a ReID dataset which is a different problem from MTMCT, while DukeMTMC is a dataset that contains only persons. Our contribution is to take advantage of spatio-temporal information derived from the videos, and also facilitated with the camera link information, to greatly improve the performance of MTMCT.
CityFlow contains 3.25 hours of traffic videos collected from 40 cameras across 10 intersections, spanning about 2.5 km, in a mid-sized U.S. city. Moreover, CityFlow covers a diverse set of road traffic types, including intersections, stretches of roadways, and highways. The length of the training videos is 58.43 minutes, while testing videos are 136.60 minutes in length. There are five scenarios in the CityFlow dataset, three of the scenarios are used for training, and the remaining two are used for testing. In total, the dataset contains 229,680 bounding boxes for 666 distinct annotated vehicle identities. The resolution is at least 960p and the frame rate is 10 FPS. Moreover, the license plates are blocked out in advance due to the privacy issues, i.e., the license plate information is not allowed to be used in the MTMCT.

In terms of implementation details, we use the dataset of the AI City Challenge 2018 \cite{naphade20182018} for training the TNT in the SCT. Since there are over 3.3K vehicles in the AI City Challenge 2018 dataset, which contains much richer information than the training set in the benchmark dataset. For both feature extraction of single camera ReID and cross camera ReID, we use ResNet-50 as the backbone network, trained with the combination of BS Triplet loss and Xent loss. In TNT, the dimension of the appearance feature is 512, the time window size is 64, and the batch size is 32. The optimizer used for training TrackletNet is Adam and the learning rate is from $10^{-3}$ to $10^{-5}$ for decreasing 10 times in every 2000 steps.

In terms of the single and cross camera ReID, the temporal attention model is trained with tracklet features from ground truth image ReID features. Same as UWIPL \cite{hsu2019multi}, we select the 4 as the clip length for each tracklet. The learning rate is $3 \times 10^{-4}$, the weight decay is $5 \times 10^{-4}$, the batch size is 32, and the network is totally trained for 800 epochs. The similarity estimation of the ReID features is calculated by the Euclidean distance. The input images are resized to $224 \times 224$. We use ResNet-50 pre-trained on ImageNet as the backbone for our model.

For MTMCT, we use IDF1, IDP, and IDR as evaluation metrics. IDF1 \cite{ristani2016performance} is used to rank the performance of each team in the CityFlow dataset. It calculates the ratio of correctly identified detections over the average number of ground truth and computed detections. More specifically, false negative ID (IDFN), true negative ID (IDTN) and true positive ID (IDTP) counts are all used to compute the identification precision (IDP), the identification recall (IDR), and the corresponding F1 score IDF1.
\begin{equation}
\begin{aligned}
    IDP &= \frac{IDTP}{IDTP+IDFP}, \\
    IDR &= \frac{IDTP}{IDTP+IDFN}, \\    
    IDF1 &= \frac{2IDTP}{2IDTP+IDFP+IDFN}.
\end{aligned}
\end{equation}
The definition of IDFN, IDTN and IDTP are as following,
\begin{equation}
    \begin{aligned}
    IDFN &= \sum_{\tau} \sum_{t \in T_\tau} m(\tau, \gamma_{m}(\tau), t, \Delta), \\
    IDFP &= \sum_{\gamma} \sum_{t \in T_\gamma} m(\tau_{m}(\gamma), \gamma, t, \Delta), \\
    IDTP &= \sum_{\tau} \text{len}(\tau)- IDFN = \sum_{\gamma} \text{len}(\gamma) - IDFP. 
    \end{aligned}
\end{equation}
where $\tau$ is the ground truth trajectory, $\gamma_m (\tau)$ means the computed trajectory that best matches $\tau$; $\gamma$ represents computed trajectory; $\tau_m (\gamma)$ is ground truth trajectory that best matches $\gamma$; $t$ is the frame index; $\Delta$ means the IOU threshold that judges whether computed bounding box matches the ground truth bounding box (here we set $\Delta=0.5$); $m(\cdot)$ is a mismatch function which is equal to $1$ if there is a mismatch at $t$; otherwise, $m(\cdot)$ is $0$.

\begin{figure*}[t]
    \centering
    \includegraphics[width=1\linewidth]{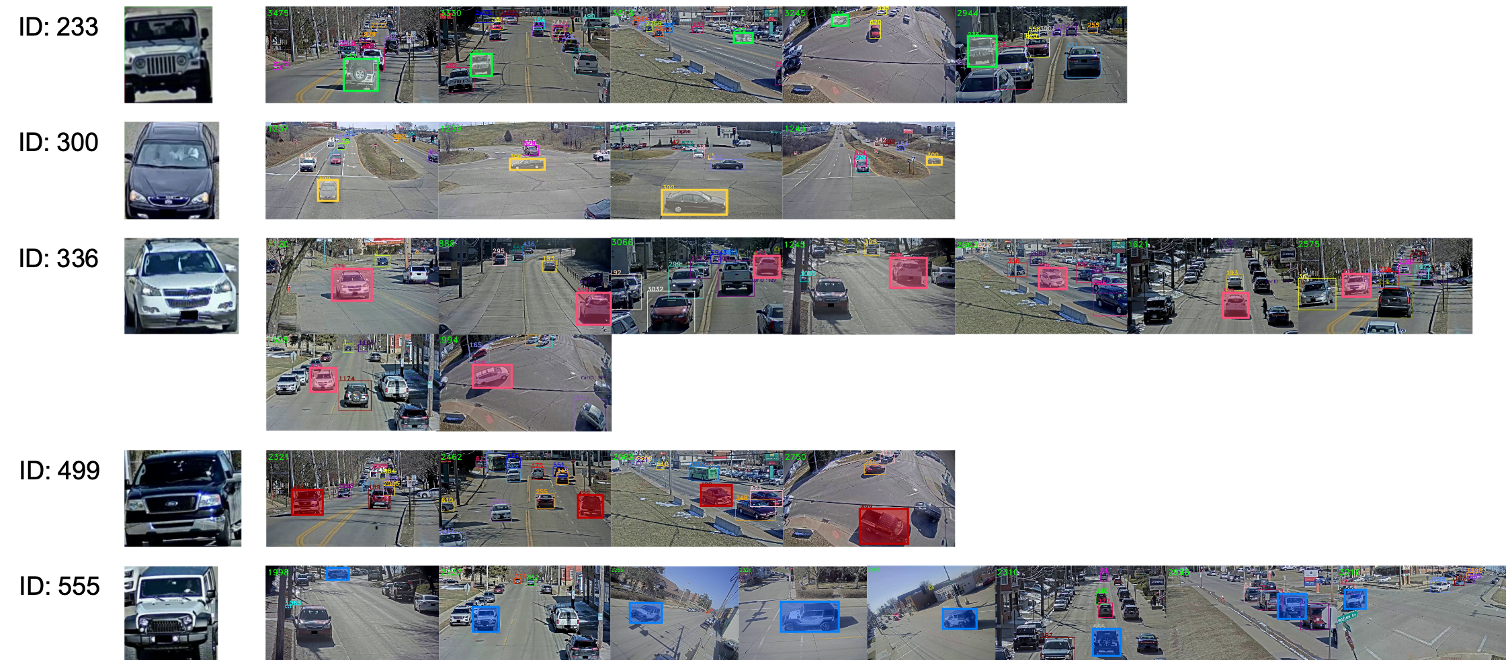}
    \caption{Qualitative results of our MTMCT results on CityFlow dataset.}
    \label{fig:qual_results}
\end{figure*}

\subsection{MTMCT Results on CityFlow}

In the CityFlow dataset, the spatio-temporal information is useful in improving the performance \cite{hsu2019multi,li2019spatio,he2019multi}. Especially, the camera links are applied to achieve the best performance \cite{hsu2019multi}. Moreover, some systems utilize data association graph for MTMCT \cite{hou2019locality} . Most of the existing MTMCT methods are based on the tracking-by-detection scheme and adapted for multi-camera views. However, the proposed method also considers traffic scenarios as well as the camera links to enhance the MTMCT performance. 

The qualitative results are shown in Fig.~\ref{fig:qual_results}, which shows that our method is generalized well for different cameras and vehicles.
Table~\ref{tab:res1} compares our methods with the state-of-the-art approaches on the CityFlow benchmark. Here, locality aware appearance metric (LAAM) \cite{hou2019locality1} is the state-of-the-art approach on DukeMTMC dataset. LAAM improves the performance of DeepCC by training the model based on both intra-camera and inter-camera metrics. The ReID features in DeepCC are extracted by DenseNet-121 with softmax and triplet loss. In terms of LAAM, they use ResNet-50 pre-trained on ImageNet. The tracklet length is set to 10 frames, and the temporal window sizes for SCT and ICT are set to 500 frames and 2400 frames. According to the experimental results, our proposed method outperforms all the state-of-the-art methods,  by incorporating the traffic-aware zones for isolated trajectories ReID and also taking advantage of the generated camera link model. Finally, we achieve IDF1 of 74.93\% for MTMCT on CityFlow. Moreover, comparing to \cite{hsu2019multi}, our system can automatically generate camera links instead of human labeling.

\begin{table}[t]
    \centering
    \captionof{table}{MTMCT results comparison on CityFlow dataset.}
    \begin{tabular}{L{3cm} C{1.5cm}}
    \hline
    Methods  & IDF1 \\
    \hline
    MOANA+BA \cite{tang2019cityflow}  & 0.3950  \\
    DeepSORT+BS \cite{tang2019cityflow}  & 0.4140  \\
    TC+BA \cite{tang2019cityflow}  & 0.4630  \\
    ZeroOne \cite{tan2019multi} & 0.5987 \\
    DeepCC \cite{ristani2018features} & 0.5660  \\
    LAAM \cite{hou2019locality1} & 0.6300  \\
    ANU \cite{hou2019locality}  & 0.6519   \\
    TrafficBrain \cite{he2019multi} & 0.6653 \\
    DDashcam \cite{li2019spatio} & 0.6865   \\
    UWIPL\cite{hsu2019multi}   & 0.7059 \\
    \hline
    \textbf{Ours}  & \textbf{0.7493} \\
    \hline
    \end{tabular}
    \label{tab:res1}
\end{table}

% \begin{table}
% \centering
% \caption{}
% \begin{tabular}{l|c c c}
% \hline 
%  & IDF1   & IDP     & IDR    \\
% \hline 
% TNT+TP  & 0.1583 & 0.4418  & 0.0959  \\
% \hline 
% TNT+TA  & 0.5237 & 0.6816  & 0.4221  \\
% \hline 
% TNT+TP+CLM   & 0.5776  & 0.5918 & 0.5779  \\
% \hline 
% TNT+TA+CLM   & 0.7059  & 0.6912 & \textbf{0.7211}  \\
% \hline 
% TSCT+TA+CLM  & \textbf{0.7493}  & \textbf{0.8071} & 0.6918 \\
% \hline 
% \end{tabular}
% \end{table}

\subsection{Ablation Studies}

The ablation studies of MTMCT is showed in Table~\ref{tab:res2}, including the combinations of TNT, TSCT, TA, and CLM. The experimental results show that each proposed component helps enhance the robustness. First, we show all modules of the proposed method are necessary. When replacing TSCT with the TNT, IDF1 based on the TA feature drops by 4.3\% on MTMCT. A similar but greater accuracy drop can be observed when the TA is replaced by the average pooling. The drop is consistent when using the same tracklet based ReID features. These results show that both the TSCT and TA are necessary components in our system. Second, from the ablation studies, the removal of the CLM causes a greater accuracy drop. The reason is the transition time constraint between cameras improves data association in tracking, so the global matching in ReID is much better in ICT. In the vehicle ReID, high inter-class similarity leads to small appearance variance of the targets. Consequently, after applying the CLM, the MTMCT performance can be largely improved. Then, we show that the TA and CLM are both important, IDF1 can be improved from 15.83\% to over 50\% by using TA or CLM. Therefore, CLM can improve MTMCT by the general ReID model.

\begin{table}[t]
\centering
\caption{The MTMCT performance for different combinations of the proposed method.}
\begin{tabular}{C{0.6cm} C{0.6cm} C{0.6cm} C{0.6cm}|C{1cm} C{1cm} C{1cm}}
\hline 
TNT & TSCT & TA & CLM & IDF1 & IDP & IDR  \\
\hline 
\checkmark & & & & 0.1583 & 0.4418  & 0.0959  \\
\checkmark & & \checkmark & & 0.5237 & 0.6816  & 0.4221  \\
\checkmark & & & \checkmark & 0.5776  & 0.5918 & 0.5779  \\
\checkmark & & \checkmark & \checkmark & 0.7059  & 0.6912 & \textbf{0.7211}  \\
\checkmark & \checkmark & \checkmark & \checkmark & \textbf{0.7493}  & \textbf{0.8071} & 0.6918 \\
\hline 
\label{tab:res2}
\end{tabular}
\end{table}

In Table~\ref{tab:res3}, we show the performance of SCT, which favorably compares with the state-of-the-art methods for SCT \cite{wojke2017simple,tang2018single,tang2019moana}. DeepSORT \cite{wojke2017simple} is an online method incorporating  Kalman-filter-based tracking and the Hungarian algorithm. TC \cite{tang2018single} is an offline method using tracklet clustering which is the winner of the AI City Challenge at CVPR 2018 \cite{naphade20182018}. MOANA \cite{tang2019moana} is a state-of-the-art approach on the MOTChallenge 2015 3D benchmark. There are three sets of available detection results, i.e., SSD512 \cite{liu2016ssd}, YOLOv3 \cite{redmon2018yolov3} and Faster R-CNN \cite{ren2015faster}, which are provided by the CityFlow dataset. According to \cite{tang2019cityflow}, SSD512 \cite{liu2016ssd} performs the best and is used by us for comparison. 
The metrics of SCT include IDF1, Multiple Object Tracking Accuracy (MOTA), Multiple Object Tracking Precision (MOTP), Recall and the mostly tracked targets (MT). According to the experimental results, the proposed TSCT method achieves the best performance.

\begin{table}[t]
\centering
\caption{SCT results on CityFlow. }
\begin{tabular}{L{2cm}|C{0.8cm} C{0.8cm} C{0.8cm} C{0.8cm} C{0.8cm}}
\hline 
Methods & IDF1  & MOTA   & MOTP   & Recall & MT   \\
\hline 
DeepSORT \cite{wojke2017simple}   & 79.5\% & 68.9\% & 65.5\% & 69.2\% & 756  \\
TC \cite{tang2018single}   & 79.7\% & 70.3\% & 65.6\% & 70.4\% & 895  \\
MOANA \cite{tang2019moana}   & 72.8\% & 67.0\% & 65.9\% & 68.0\% & 980  \\
\hline 
\textbf{Ours} & \textbf{88.4\%} & \textbf{79.3\%} & \textbf{75.2\%} & \textbf{85.1\%} & \textbf{1726} \\
\hline 
\end{tabular}
\label{tab:res3}
\end{table}

\section{Conclusion}
\label{sec:conclusion}

In this paper, we propose a novel approach for multi-target multi-camera tracking (MTMCT) of vehicles, which includes traffic-aware single camera tracking (TSCT), trajectory-based camera link model (CLM), and vehicle re-identification (ReID). 
Finally, hierarchical clustering is utilized to merge the vehicle trajectories from different cameras and generates the MTMCT results.
From our experiments, the proposed method is shown to be effective and robust. It also achieves a new state-of-the-art performance on the CityFlow dataset.

\section*{Acknowledgement}

This work was partially supported by Electronics and Telecommunications Research Institute (ETRI) grant funded by the Korean government. [20ZD1100, Development of ICT Convergence Technology for Daegu-GyeongBuk Regional Industry]

\bibliographystyle{ACM-Reference-Format}
{\balance \bibliography{sample-base}}

\end{document}